%% file: chronos.tex
\documentclass[sigconf]{acmart}

\setcopyright{acmlicensed}
\acmConference[]{}{}{} 
\acmYear{2019}
\copyrightyear{2019}

\acmPrice{15.00}
\acmDOI{10.1145/1122445.1122456}
\acmISBN{978-1-4503-9999-9/18/06}

\usepackage[utf8]{inputenc}
\usepackage[T1]{fontenc} 
\usepackage{nicefrac} 
\usepackage{microtype} 
\usepackage{verbatim}
\usepackage{multirow}
\usepackage{booktabs} 
\usepackage{amsfonts}
\usepackage{algorithm, algorithmicx, algpseudocode}
\usepackage{amsmath}
\usepackage{array}
\usepackage{subfigure}
\usepackage{balance}
\usepackage{multicol}
\usepackage{color}
\usepackage{epsfig}
\usepackage{natbib}
\usepackage{xspace}
\usepackage{bm}
\usepackage{enumitem}
\usepackage{diagbox}
\usepackage{url}
\usepackage[font=small, labelfont=bf, textfont=md]{caption}
\newcommand{\hide}[1]{} 
\newcommand{\vpara}[1]{\vspace{0.3em}\noindent\textbf{#1 }}

\newcommand{\equationref}[1]{Eq~\ref{#1}}
\newcommand{\secref}[1]{Section~\ref{#1}} 
\newcommand{\alref}[1]{Algorithm~\ref{#1}} 
\newcommand{\figref}[1]{Figure~\ref{#1}} 
\newcommand{\tableref}[1]{Table~\ref{#1}} 
\newcommand{\beq}[1]{\begin{equation}\vspace{0.em}#1\vspace{0.em}\end{equation}}

\newcommand{\methodname}{Generative Mixture Nonparametric Encoder}
\newcommand{\methodshort}{GeNE}

\newcommand{\kaggletraffic}{Web Traffic Time Series Forecasting\xspace}
\newcommand{\kaggletrafficshort}{WebTraffic\xspace}
\newcommand{\trafficdataset}{Telecom Monthly Plan\xspace}
\newcommand{\trafficshort}{MonthPlan\xspace}
\newcommand{\flowdataset}{Information Networks Supervision\xspace}
\newcommand{\flowshort}{NetFlow\xspace}
\newcommand{\timedataset}{Watt-hour Meter Clock Error\xspace}
\newcommand{\timeshort}{ClockErr\xspace}

\hide{

}

\begin{document}
\title{Capturing Evolution Genes for Time Series Data}

\author{Wenjie Hu$^{\dagger}$, Jianping Huang$^{\ast}$, Liang Wu$^{\ast}$, Yang Yang$^{\dagger}$, Zongtao Liu$^{\dagger}$, Zhanlin Sun$^{\dagger}$, Bingshen Yao$^{\S}$, Ke Chen$^{\ast}$}

\affiliation{
	\institution{$^{\dagger}$College of Computer Science and Technology, Zhejiang University, Hangzhou, China}
	\institution{$^{\ast}$State Grid Zhejiang Electric Power Supply Co. Ltd., China}
	\institution{$^{\S}$Rensselaer Polytechnic Institute, Troy, New York, USA}
	\institution{$^{\dagger}$\{aston2une, yangya, tomstream, zhanlinsun15\}@zju.edu.cn, $^{\ast}$\{huang\_jianping, wuliang1127, chenke\}@zj.sgcc.com.cn, $^{\S}$yaob@rpi.edu}
}

\renewcommand{\shortauthors}{Hu et al.}

\input{abstract.tex}


\maketitle
\input{intro.tex}
\input{setup.tex}
\input{model.tex}
\input{exp.tex}

\input{interpreter.tex}
\input{related.tex}

\input{conclusion.tex}

\balance
\bibliographystyle{ACM-Reference-Format}
\bibliography{reference}


\end{document}

%% file: abstract.tex

\begin{abstract}
The modeling of time series data is becoming increasingly critical in a wide variety of applications.
Overall, data evolves by following different patterns, which are generally caused by different user behaviors. 
Given a time series, we define the \textit{evolution gene} to capture the latent user behaviors and to describe how 
the behaviors lead to the generation of time series. 
In particular, we propose a uniform framework that 
recognizes different evolution genes of segments by learning a classifier,  and adopt an adversarial generator to implement the evolution gene by estimating the segments' distribution.
Experimental results based on a synthetic dataset and five real-world datasets show that our approach can not only achieve a good prediction results (e.g., averagely +$10.56\%$ in terms of F1), but is also able to provide explanations of the results. 
\end{abstract}

%% file: intro.tex

\section{Introduction}
\label{sec:intro}

\begin{figure}[t]
	\begin{minipage}{0.5\textwidth}
		\centering
		\includegraphics[width=1.0\textwidth]{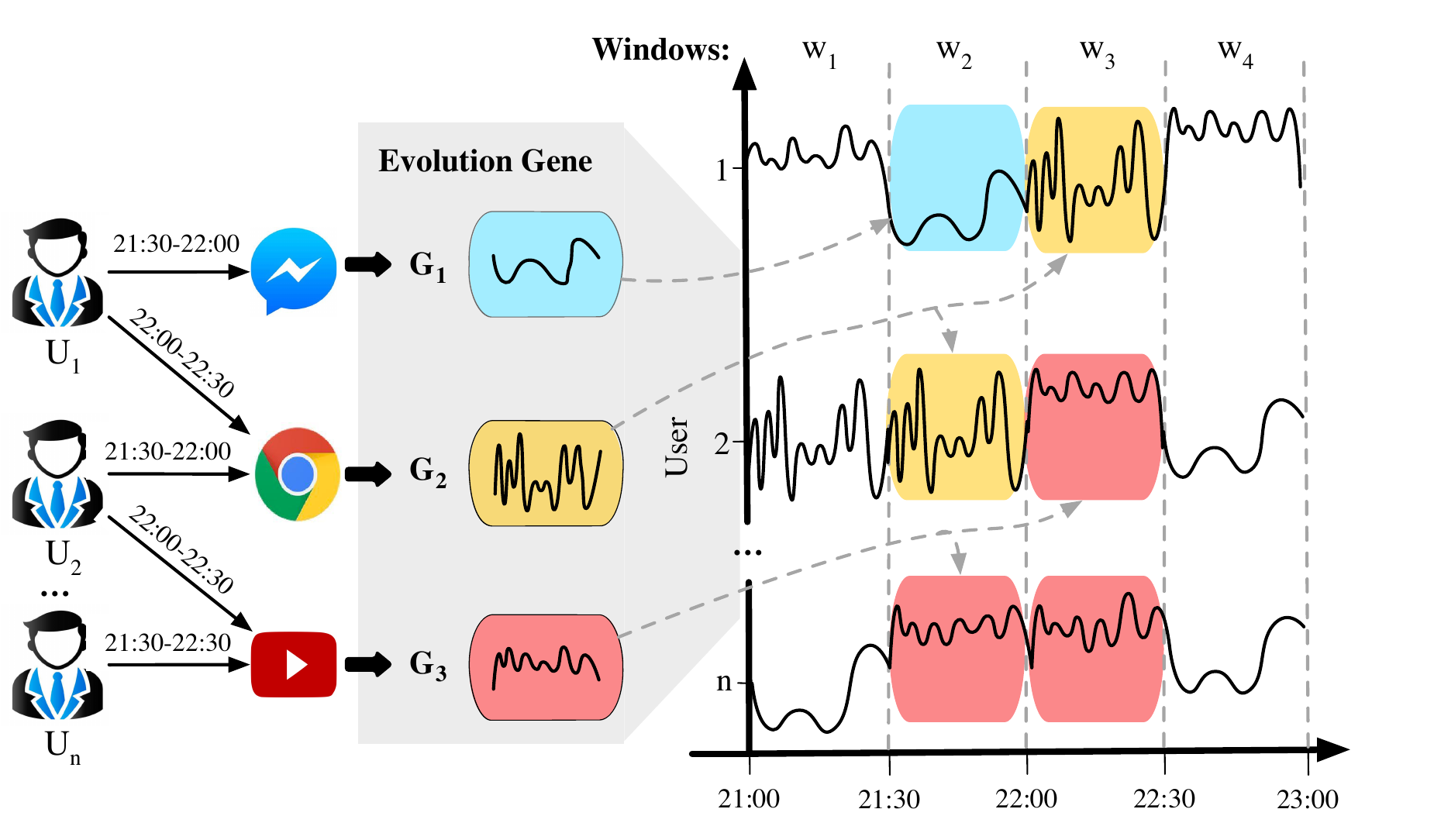}
		\caption{
			An illustration of genes when modeling network flow. Each user action (e.g., chatting, watching a movie, browsing web-pages) has a gene to generate corresponding segments of time series. 
		}
		\label{fig:example}
	\end{minipage}
\end{figure}

The modeling of time series data has attracted significant attention from the research community in recent years, due to its broad applications in different domains such as financial marketing and bioinformation\citep{janakiraman2017finding, Du2016Recurrent, Barbosa2016Averaging,chapfuwa2018adversarial}. 
For example, a communication company might formulate a user's network flow as a time-sensitive segments $\{\mathcal{X}_1, \mathcal{X}_2, \cdots, \mathcal{X}_n\}$, where each element $\mathcal{X}_i$ denotes how the user uses flow at the $i$-th time window. 
The systems then work to understand the user's behavior behind each segment, and predict his or her flow cost $\mathcal{X}_{n+1}$ in future. 
Appropriate phone plans are then recommended to the user on the basis of this model. 
More specifically, users' flow-costs evolve over time by following different patterns.  
As \figref{fig:example} illustrates, when watching movies, a user begins to use a certain volume of flow in a certain time period, but uses a low flow during the chat.
Meanwhile, another user has unstable flow loads when surfing the Internet, that flow will be higher when clicking pages and lower when reading pages. 

Different evolution patterns of time series reflect different user behaviors, which exist a certain regularity. 
For example, users usually browse the Internet for some information after chatting, or spend a long time watching a movie, and occasionally cut out because of chatting.
Thus, if a method is able to extract user behaviors behind given segments, learn how each behavior leads the generation of segment, 
and capture the transition of user behaviors, 
it can be more predictive in time series. 
However, to the best of our knowledge, most existing works, such as deep neural network-based models (e.g., LSTM and VAE)~\citep{Du2016Recurrent, Kingma2013Auto} do not distinguish different patterns and use only one single model for generating all data. 
Meanwhile, traditional mixture models (e.g. GMM and HMM)~\citep{bouttefroy2010on, Yang2014HMM} ignore the transition of user behaviors over time which turns out to have good performances in recent research. 

\vpara{Evolution gene.}
In this paper, we propose the concept of \textit{evolution gene} (or \textit{gene} for short) to quantitatively describe how each kind of user behavior generates the corresponding time series.
More specifically, we define the \textit{gene} $\mathbf{G}$ as a generative model that captures the distribution patterns and learns to generate the segments. 
As shown in \figref{fig:example}, there are three different genes, each corresponding to a particular user behavior. 
For instance, $\mathbf{G}_1$ generates the flow segments of chatting online, while $\mathbf{G}_3$ generates the flow segments of watching movies.  
For a given sequence of time series segments $\{\mathcal{X}_1, \cdots, \mathcal{X}_n\}$, we aim to learn and extract the gene $\mathbf{G}_k$ of each segment $\mathcal{X}_n$, based on which we further predict the future value $\mathcal{X}_{n+1}$ and the event that will happen at the time window $n+1$. 

This problem is nontrivial. 
A straightforward baseline is to first cluster these segments, assign each cluster a gene, and then learn the generator for each cluster independently. 
However, other than considering the distance of samples like most clustering algorithms do, our goal is to determine which segments share similar distribution and sequential patterns. 
Therefore, the above baseline does not work well, as will be demonstrated in our experiments 
(see \tableref{tb:exp:assign}). 
The question of how to design an appropriate algorithm to recognize genes is the major challenge in this work.

Once aware of time series genes, 
we then aim to estimate what event will happen in future. 
Traditional works mainly predict events according to the data value of a snapshot, such as dynamic time warping \citep{Lines2015Time}, complexity-invariant distance\citep{Batista2014CID} and elastic ensemble\citep{Lines2015Time}. They concentrate on different distance measurements and find the nearest sample. 
However, the behaviors' evolution are more important for the prediction task. 
For example, an watt-hour meter experiencing a sudden drop in electricity consumption implies an abnormal event, which may be either caused by early damage to the meter or power-stealing behavior. 
Building the connection between the behavior evolution and the future event is another challenge. 

Here, we propose a novel model: \underline{G}enerative Mixtur\underline{e} \underline{N}onparam-etric \underline{E}ncoder (\methodshort), which distinguishes the distribution patterns of time series by learning generating the corresponding segments. This model has three major components:  \textit{gene recognition}, aims at learning the corresponding genes of segments; \textit{gene generation}, aims at learning generating segments from each gene; \textit{gene application}, aims at modeling the behavior evolution and applying the learned genes to future value and event prediction. 

We evaluate the proposed model on a synthetic dataset and four real-world datasets.
The experimental results demonstrate our advantage over several state-of-the-art algorithms on three different tasks (e.g., averagely +$10.56\%$ in terms of F1). 
Moreover, we demonstrate some meaningful interpretation of our method by visualizing the behavior evolution. 
We apply our method to predict clock error fluctuation of watt-hour meter in the State Grid of China\footnote{The state-owned electric utility of China, and the largest utility company in the world.} and help to reduce electrical equipment maintenance workloads by 50\%, which cost around \$300 million per year\footnote{http://www.sgcc.com.cn/ywlm/index.shtml}.

Accordingly, our contributions are as follows: 
\begin{itemize} [leftmargin=*]
\item We define the concept of evolution gene to formally describe how latent behaviors generate time series;
\item We propose a novel and uniform framework that distinguishes the 
latent behaviors and model their evolution on time series.
\item We construct sufficient experiments,  based on both synthetic and real-world datasets, to validate whether our method is capable of 
modeling time series.
Experimental results exhibit our method's advantage over eleven state-of-the-art algorithms in different prediction tasks.
\item We have deployed our model to the real scenario for identifying abnormal watt-hour meters, under the corporation with State Grid of China. Through the application, we find that the genes learned by our model can provide some explanations for the anomalies in practice.
\end{itemize}

%% file: setup.tex

\section{\methodname}
\label{sec:model}
\subsection{Preliminaries}
The task considered in this paper is to capture the behavior evolution behind time series, and then to utilize these patterns to predict the value and event that will happen in the future. 

Formally, let $\mathcal{X} \in \mathbb{R}^{N \times T \times S}$ be an observation-sequence with $N$ time windows in a time series data.
Each $\mathcal{X}_{n} = \{x_t\}_{t=1}^T \in \mathbb{R}^{T \times S}$ is a segment in the time window whose length is $T$. $T$ has a physical meaning, such as one day has 24 hours or one month has 30 days.
Each $x_t \in \mathcal{X}_{n} $ is a single- or multi-variate observation with $S$ variables, denoted as $x_t = \{x_{t}^{(s)}\}_{s=1}^S \in \mathbb{R}^{S}$. 
$Y = \pi \in \Pi$ represents the future event occurring under observation-sequence $\mathcal{X}$, where $\Pi \subset \mathbb{Z}$ is the set of markers and $\pi$ is the specific event marker.
We define $\mathcal{A}_n \in \mathbb{R}^K$ as the recognition probability of $\mathcal{X}_n$ for $K$ behaviors, where $0\le \mathcal{A}_n^{(k)} <1$ and $\sum_{k=1}^{K}\mathcal{A}_n^{(k)} =1$.
We aim to infer the future values $\mathcal{X}_{(N+1)}$ and event probability $\mathbf{P}(\mathcal{Y} | \mathcal{X, A}) $.
Here, we propose a novel generative method to model the time series $\mathcal{X}$ that focuses on distinguishing the distribution patterns of segments and their overall behavior evolution on the time series.

%% file: model.tex

\subsection{General Description}
\begin{figure*}[t]
	\begin{minipage}{1.\textwidth}
		\centering
		\includegraphics[width=0.8\textwidth]{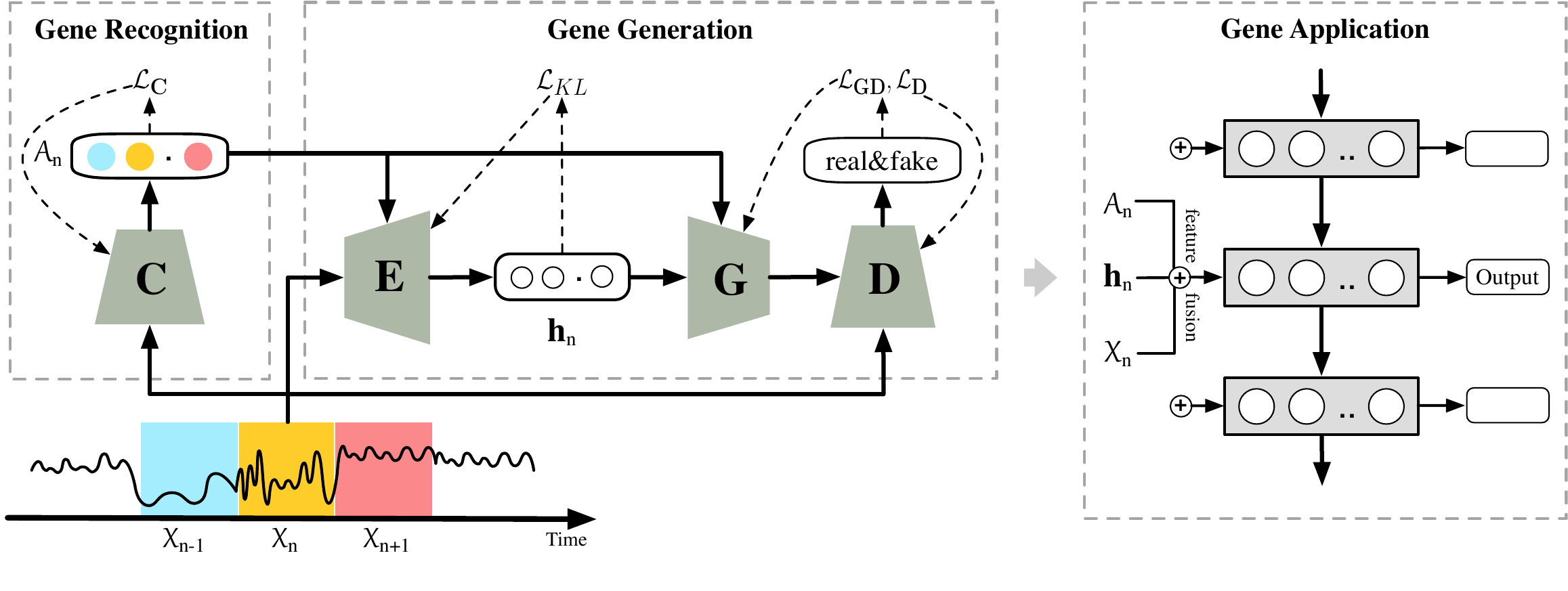}
		\caption{Structure of the proposed model, which consists of three components: gene recognition, gene generation, and gene application.}
		\label{fig:cluster}
	\end{minipage}
\end{figure*}
We propose a novel model, \underline{G}enerative Mixtur\underline{e} \underline{N}onparametric \underline{E}ncoder (\methodshort), which distinguishes different behaviors behind the time series by learning the corresponding genes, and captures distribution patterns of each segment $\mathcal{X}_n$ to make prediction.
We put these two objectives into a uniform framework. 
As \figref{fig:cluster} shows, given the number of genes $K$, the proposed model consists of three components: 
\textit{gene recognition}, aims at recognizing the corresponding genes of segments. 
\textit{gene generation}, aims at generating segments of each gene;
\textit{gene application}, aims at applying the learned genes to the downstream tasks, such as prediction or classification of time series.

\vpara{Gene recognition.} 
This component is to 
recognize the corresponding genes of each segment $\mathcal{X}_n$, 
which can be implemented in several different ways like clustering algorithms. 
In this work, for distinguishing the distribution and sequential patterns of segments simultaneously, we propose a sequence-friendly classification network $\mathbf{C}$ (implemented by RNN or LSTM) to improve the recognition from the clustering algorithms. 
We practically compare this method with other potential implementations and find that it has the best performance (see details in Table~\ref{tb:exp:assign} of Section~\ref{sec:exp}). 

\vpara{Gene generation.} 
This component is to learn genes for generating segments, which aims at capturing segments' distribution patterns.
In this work, gene generation is implemented by an adversarial generator $(\mathbf{G}|\mathbf{D})$, which structure is like CVAE-GAN~\citep{2017arXiv170310155B}, but the loss is more concise. 
It captures the superior distribution patterns which outperform other implementations (see details in \secref{sec:exp}).

\vpara{Gene application.} 
Genes recognize the behaviors behind the segments, which is represented by the different distribution patterns. 
They can be combined sequentially on the time series $\mathcal{X}$, just like the biological genetic code.
Hence, we propose a recurrent structure to combine these genes on the time series and apply them to the downstream tasks, which leads to a superior predictive and interpretive model as \secref{sec:exp} and \secref{sec:interpreter} show.

Overall, gene recognition provides the supervised information to guide the gene generation, which improves the ability of capturing segments' distribution patterns. They are irrelevant to the downstream tasks $\mathcal{Y}$ and thus can be off-line trained. 
Gene application is based on the “end-to-end” learning, which adjusts gene recognition and generation for the real-time response.
We will introduce each component in detail in the following chapters. 

\subsection{Gene Recognition}
As described in \secref{sec:intro}, time series data evolves by different distributions, which are generally caused by different behaviors. Hence, we can find these behaviors behind the time series via capturing the distributions.
However, the traditional clustering algorithms focus on the distance between different samples. They treat each variable as an independent individual without considering the sequential similarity, and thus is not suit for the gene recognition. 
We explore a novel method to overcome these difficulties mentioned above. 

Generally, given the number of genes $K$, we first initialize a recognition $\mathcal{A}^{(0)}$ via traditional distance-based clustering algorithms $f$, such as K-means, input of which is the mean and variance of each segment's variables. The formulation is:
\beq{
\begin{split}
	\mu_n &= \frac{1}{T}\sum_{t=1}^{T} \mathcal{X}_n, \\
	\mathcal{A}_n^{(0)} &= f\left(\mu_n, \frac{1}{T}\sum_{t=1}^{T} (\mathcal{X}_n - \mu_n)^2 \right)
	\label{eq:initial}
\end{split}
}
The motivation here is that, if the mean and variance are close in distance, the segments are more likely to have a similar distribution \citep{Bagnall2017}, thus they should be recognized into the same gene.

However, there may be two segments with different sequential patterns but similar distribution, such as the trend, mutations, or zero numbers etc.
Therefore, we need a method to distinguish these sequential patterns for recognizing genes.
Following this idea, we design a sequence-friendly classification network $\mathbf{C}(\mathcal{X}_n; \theta_{\mathbf{C}})$, 
where $\theta$ is the model parameters, to capture the sequential patterns in segments and improve the quality of the current gene recognition.
Specifically, the network $\mathbf{C}$ takes raw segments $\mathcal{X}_n$ as input and outputs a $K$-dimensional vector, and then turns into probabilities using a softmax function. The output of each entry represents the probability $\mathbf{P}(k|\mathcal{X}_n)$. In the training stage, the deep neural network $\mathbf{C}$ tries to minimize the cross-entropy loss as follow:
\beq{
\mathcal{L}_{\mathbf{C}}= -\mathbb{E}_{\mathcal{X}\sim p_r}[\log \mathbf{P}(k |\mathcal{X}_n)]
\label{eq:classifier}
}
where $p_r$ is the real empirical joint distribution of segments, which can be estimated by sampling.
We take the network $\mathbf{C}$'s recognition as the newly recognition and repeat the steps until the error rate $\frac{|\mathcal{A} \ominus \mathcal{A}'|}{|\mathcal{A}|}$ converged, where $\mathcal{A}$ and  $\mathcal{A}'$ are the old and new gene recognition at each iteration.
For the implementation of the classification network $\mathbf{C}$, we use RNN or a modern variant like LSTM, which is good at capturing the sequential patterns in the time series.

\subsection{Gene Generation}
The segments corresponding to the same gene have the similar distributions, that the non-parametric generative model is a natural and effective way to estimate them. 
As \figref{fig:cluster} shows, we input segments with the gene recognition into a CVAE-GAN structure, 
which encode the segments into the hidden space under the condition of gene recognition, and discriminate the fake samples generated from the variational approach.

More specifically, for each segment $\mathcal{X}_n$ and its gene recognition $\mathcal{A}_n$, 
each gene represents its distribution patterns by an encoder network $\mathbf{E}(\mathcal{X}_n, \mathcal{A}_n; \theta_{\mathbf{C}})$, which obtains a mapping from the real segment $\mathcal{X}_n$ to the hidden vector $\mathbf{h}_n$.
We use a multivariate Gaussian distribution with a diagonal covariance structure to present the variational approximate posterior:
\beq{
\label{eq:vaedefinition}
\log \mathbf{E}(\mathbf{h}_n | \mathcal{X}_n, \mathcal{A}_n) = \log \mathcal{N}(\mathbf{h}_n; \bm{\mu},\bm{\delta}^2\mathbf{I}, \mathcal{A}_n)
}
Based on the variational approach, for each segment, when the encoder network $\mathbf{E}$ outputs the mean $\bm{\mu}$ and covariance $\bm{\delta}$ of the hidden vector, genes can sample the hidden vector $\mathbf{h}_n = \bm{\mu} + z \odot \exp (\bm{\delta})$,  where $z \sim \mathcal{N}(0, \mathbf{I})$ is a random vector and $\odot$ represents the element-wise multiplication. We use the KL loss to reduce the gap between the prior $\mathbf{P}(\mathbf{h}_n)$ and the proposal distributions, i.e:
\beq{
\label{eq:likelihoods}
\mathcal{L}_{KL} = \frac{1}{2}(\bm{\mu}^T\bm{\mu} + \sum (\exp(\bm{\delta}) - \bm{\delta} - 1))
}
After obtaining the mapping from $\mathcal{X}_n$ to $\mathbf{h}_n$, each gene can then map the generated segments by an generator network, which formulates as $\mathcal{X}'_n = \mathbf{G}_k(\mathbf{h}_n, \mathcal{A}_n; \theta_{\mathbf{G}})$.
The discriminator network $\mathbf{D}(\mathcal{X}_n; \theta_{\mathbf{D}})$ estimates the probability that a segment comes from the real samples rather than $\mathcal{X}'_n$, which tries to minimize the loss function:
\beq{
	\mathcal{L}_{\mathbf{D}} = -\mathbb{E}_{\mathcal{X}\sim p_r}\left[\log \mathbf{D}(\mathcal{X}_n)\right] - 
	\mathbb{E}_{\mathbf{h} \sim p_z}\left[\log(1-\mathbf{D}(\mathcal{X}'_n))\right]
\label{eq:discriminator}
}
where $p_r$ is the real empirical joint distribution and $p_z$ is a simple distribution, e.g., isotropic Gaussian or uniform.
The training procedure for $\mathbf{G}_k$ is to maximize the probability of $\mathbf{D}$ making a mistake, while $\mathbf{G}_k$ tries to minimize:
\beq{
	\mathcal{L}'_{\mathbf{G}_k\mathbf{D}} = -\mathbb{E}_{\mathbf{h} \sim p_z}\left[\log(\mathbf{D}(\mathcal{X}'_n))\right]
}

In practice, the distributions of “real” and “fake” samples may not overlap with each other, especially at the early stage of the training process. Hence, the discriminator network $\mathbf{D}$ can separate them perfectly, that is, we always have $\mathbf{D}(\mathcal{X}_n) \rightarrow 1$ and $\mathbf{D}(\mathcal{X}'_n) \rightarrow 0$. 
Therefore, when updating genes $\mathbf{G}$, the gradient $\partial \mathcal{L}'_{\mathbf{GD}} / \partial \mathbf{D}(\mathcal{X}'_n) \rightarrow -\infty$. Consequently, the training process of $\mathbf{G}$ will be unstable. Recent works \citep{Arjovsky2017Wasserstein} also theoretically show that training GAN often involves dealing with the unstable gradient of $\mathbf{G}$.

To solve this problem, we use a mean feature matching objective for the gene. The objective requires the center features of the generated samples to match the center features of the real samples. Let $\mathcal{F}_{\mathbf{D}}(\mathcal{X}_n)$ denote features on an intermediate layer of the discriminator network.  Then $\mathbf{G}_k$ tries to minimize the loss function:
\beq{
	\mathcal{L}_{\mathbf{G}_k\mathbf{D}} = ||~ \mathbb{E}_{X \sim p_r} \mathcal{F}_{\mathbf{D}}(\mathcal{X}_n) -\mathbb{E}_{\mathbf{h} \sim p_z}\mathcal{F}_{\mathbf{D}}(\mathcal{X}'_n) ~||^2_2 
	\label{eq:generator}
}
In order to maintain simple in our experiment, we choose the input of the last fully connected layer of network $\mathbf{D}$ as the feature $\mathcal{F}_{\mathbf{D}}$. Both the $\mathbf{G}$ and $\mathbf{D}$ are trained by a stochastic gradient descent (SGD) optimization algorithm. 

We present the procedure of \methodname~ in \alref{alg:gene}.

\begin{figure}[htb]
	\centering
	\begin{minipage}{1.0\linewidth}
		\begin{algorithm}[H]
			\renewcommand{\algorithmicrequire}{\textbf{Input:}}
			\renewcommand{\algorithmicensure}{\textbf{Output:}}
			\caption{\textbf{~Off-line training of \methodshort}}
			\label{alg:gene}
			\begin{algorithmic}[1]
				\Require time series $\mathcal{X} \in \mathbb{R}^{N \times T \times S}$, number of gene $K$
				\Ensure evolution genes $(\mathcal{A}, \mathbf{h})$
				
				\Procedure{OuterGenes}{$\mathcal{X}$} 
				\State $\mathcal{A}^{(0)} \in \mathbb{R}^{N \times K} \gets$ compute the initial recognition as \equationref{eq:initial}
				\While{the error rate $\frac{|\mathcal{A} \ominus \mathcal{A}'|}{|\mathcal{A}|}$ has not converged}
				\State $\mathcal{A} = \mathcal{A}'$
				\While{$\theta_{\mathbf{D}}$, $\theta_{\mathbf{G}}$ and $\theta_{\mathbf{C}}$ have not converged}
				\State Sample $\{\mathcal{X}_n^{(m)}, \mathcal{A}_n^{(m)}\}_{m=1}^M \sim p_r$ a batch from real segments and gene recognition $\mathcal{A}_n$
				\State $\theta_{\mathbf{C}} \leftarrow \nabla_\theta[\frac{1}{M}\sum_{m=1}^{M} \mathcal{L}_{\mathbf{C}}^{(m)}]$ 
				\State $\mathcal{A}' = \arg \max \mathbf{C}(\mathcal{X})$ 
				\For{$k \gets \arg \max(\mathcal{A}')$}
				\State  $\theta_{\mathbf{D}} \leftarrow \nabla_\theta[\frac{1}{M}\sum_{m=1}^{M} \mathcal{L}_D^{(m)}]$
				\State $\theta_{\mathbf{\mathbf{E}}}, ~ \theta_{\mathbf{\mathbf{G}_i}} \leftarrow \nabla_\theta[\frac{1}{M}\sum_{m=1}^{M} (\mathcal{L}_{KL}^{(m)} + \mathcal{L}_{GD}^{(m)})]$
				\EndFor
				\EndWhile
				\EndWhile
				\State $\mathbf{h} \gets \mathbf{E}(\mathcal{X}, \mathcal{A})$
				\EndProcedure	
			\end{algorithmic}  
		\end{algorithm}
	\end{minipage}
\end{figure}

\subsection{Gene Application and Learning}
Genes recognize the behaviors behind the segments, which is represented by the different distribution patterns. 
They can be combined sequentially on the time series. The sequence of genes reveals the behavior evolution of this time series, which leads to a superior predictive and interpretive model (Section~\ref{sec:interpreter} will present it in detail).
In this work, we propose a recurrent structure to combine these genes on the time series and apply them to the downstream tasks, which mainly focus on the prediction and classification of time series.

Formally, given observation-sequence $\mathcal{X} \in \mathbb{R}^{N \times T \times S}$, we first get all the gene recognition $\mathcal{A}$ by network $\mathbf{C}$, and the 
distribution patterns $\mathbf{h}$ of the most likely genes. We fuse these features using a hybrid RNN structure, as shown in \figref{fig:cluster}, which the latent vector is donated as $\mathbf{H}$.

\vpara{Feature Fusion.} We update the latent vector $\mathbf{H}_n$ after receiving the memory $\mathbf{H}_{n-1}$ from the past, segment $\mathcal{X}_n$, gene recognition $\mathcal{A}_n$, and genes' 
patterns $\mathbf{h}_n$. The formulation is:
\begin{equation}
	\mathbf{H}_n = tanh(W\cdot(\mathcal{X}_n;\mathcal{A}_n;\mathbf{h}_n) + U\cdot\mathbf{H}_{n-1} + b)
	 \label{eq:rnn_hidden}
\end{equation}
where $W$, $U$ and $b$ are the learnable weight or bias vectors, and $\cdot$ is the matrix product.

\vpara{Output}
The last application layer apply an ``end-to-end" mechanism to the downstream tasks (predicting the future value $\mathcal{X}_{N+1}$ and the event $\mathcal{Y}$). 
$\Psi$ denotes the neural networks, which takes the last latent vector $\mathbf{H}_N$ as input. 
For the value prediction, $\Psi$ outputs a vector, and then turns into predicted value using a Relu function. In the experiment, we use DCNN~\citep{zeiler2013visualizing} as $\Psi$ and back propagate mean-square loss to train the network, which the loss can be formulated as:
\beq{
	\mathcal{L}_{app} = || \mathcal{X}_{N+1} - \Psi(\mathbf{H}_{N})||_2^2\\	 
}
For the event prediction, it can be turned into a classification problem. $\Psi$ outputs a $\Pi$-dimensional vector, and then turns into probabilities using a softmax function. In the training stage, model tries to minimize the cross-entropy loss as follow:
\beq{
	\mathcal{L}_{app} = - \mathbb{E}_{\mathbf{H} \sim p_r}[\log \mathbf{P}(\mathcal{Y} = \pi | \mathbf{H}_N )]
}
Above all, we can enhance the performance of prediction by genes.

\begin{figure}[tb]
	\centering
	\begin{minipage}{1.0\linewidth}
		\begin{algorithm}[H]
			\renewcommand{\algorithmicrequire}{\textbf{Input:}}
			\renewcommand{\algorithmicensure}{\textbf{Output:}}
			\caption{\textbf{~Downstream application of \methodshort}}
			\label{alg:prediction}
			\begin{algorithmic}[1]
				\Require time series $\mathcal{X} \in \mathbb{R}^{N \times T \times S}$, event labels $\mathcal{Y}$, evolution genes $(\mathcal{A}, \mathbf{h})$
				\Ensure predicted label $\mathcal{Y}'$
				
				\Procedure{GenePredict}{$\mathcal{X}, \mathcal{A}, \mathbf{h}$}
				\While{the parameters of \methodshort have not converged}
				\State Sample $\{\mathcal{X}^{(\eta)}, \mathcal{A}^{(\eta)}, \mathbf{h}^{(\eta)}\} $ a batch from $(\mathcal{X}, \mathcal{A}, \mathbf{h})$
				\For{each time window $n \in N$} 
				\State $\mathcal{Q}_n\gets [\mathcal{Q}_{n-1}, \mathcal{X}_n, \mathcal{A}_n, \mathbf{h}_n]$
				\EndFor
				\State $\Psi(\mathcal{Q}_N) \gets$ Abstract the last output for prediction tasks
				\State $\theta_{app} \gets \nabla_\theta \left[\frac{1}{\eta}\sum_{n=1}^{\eta} (\mathcal{L})\right]$
				\State $\theta_{\mathbf{E}}, \theta_{\mathbf{G}} \gets \alpha \times \nabla_\theta \left[\frac{1}{\eta}\sum_{n=1}^{\eta} (\mathcal{L})\right]$
				
				\EndWhile
				\EndProcedure
				
			\end{algorithmic}  
		\end{algorithm}
	\end{minipage}
\end{figure}
\vpara{End-to-end learning.}
We next introduce the end-to-end learning of \methodshort. 
The complete loss $\mathcal{L}$ of \methodshort~network is as follows:
\beq{
	\mathcal{L} = \mathcal{L}_{app} + \lambda_1\left( \mathcal{L}_{\mathbf{D}} + \mathcal{L}_{\mathbf{G}_k\mathbf{D}} + \mathcal{L}_{KL}\right) + \lambda_2 \mathcal{L}_{\mathbf{C}}
}
where ${\lambda_1, \lambda_2} > 0$ are tuning parameters, which control the trade-off between 
the gene recognition and gene generation relative to the gene application objective.
In our experiments, we set $\lambda_1=\lambda_2=1$.

Intuitively, classifier $\mathbf{C}$ is trained to fit the current recognition of segments. 
Meanwhile, the elements ($\mathbf{E}, \mathbf{G}, \mathbf{D}$) of genes are trained via an adversarial process on the real/fake samples under the condition of $\mathbf{C}$'s output. 
More specifically, in each iteration, we first train $\mathbf{C}$ to output the current recognition, and then train $\mathbf{E, G, D}$ to capture the segments' distribution. 
The recognition of $\mathbf{C}$ distinguishes the segments $\mathcal{X}_n$ and gives them specific gene index $k$, so that unsupervised adversarial training is transferred to supervised adversarial training. 
It improves the ability of the gene to capture distribution patterns. 
Then, we compare the new and old recognition and determine whether to end the iteration.
For the application layer, recursive hidden vector $\mathbf{H}$ fuses these patterns transferred from gene recognition and generation, and applies them into the prediction tasks. We back propagate the loss $\mathcal{L}_{app}$ to learn the gene application and use lower learning rate to adjust gene recognition ($\mathbf{C}$) and gene generation ($\mathbf{E, G, D}$).
We present the complete procedure in \alref{alg:prediction}.

%% file: exp.tex

\section{Experiment}
\label{sec:exp}
\begin{figure*}[th]
	\centering
	\subfigure[Synthetic data.]{
		\label{fig:violin:PA}
		\includegraphics[width=0.23\textwidth]{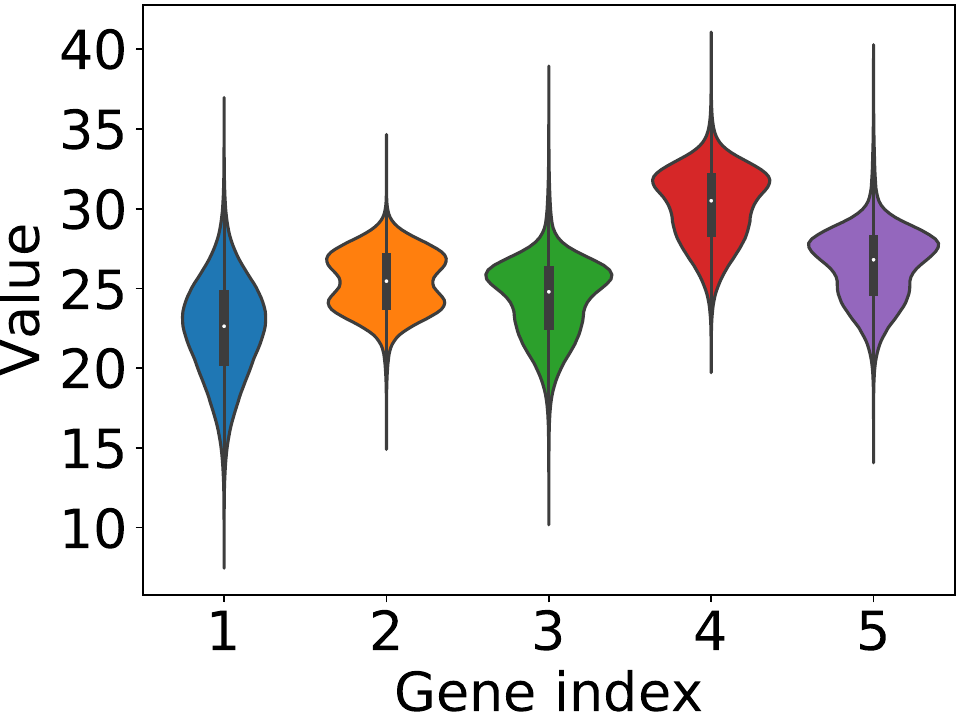}
	}
	\subfigure[CVAE]{
		\label{fig:violin:PB}
		\includegraphics[width=0.23\textwidth]{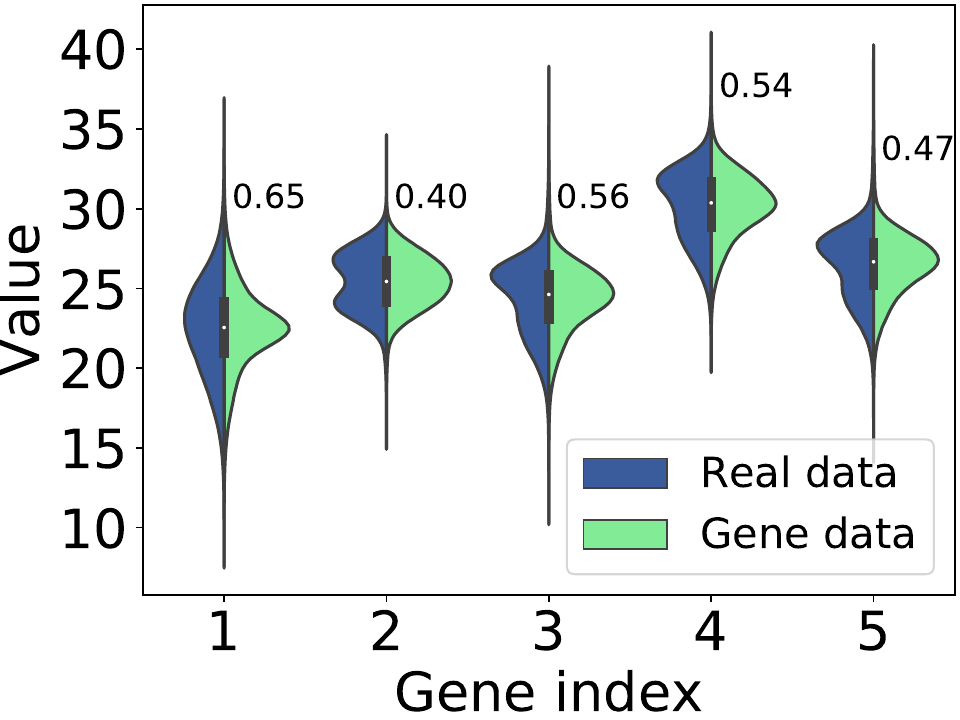}
	}
	\subfigure[CGAN]{
		\label{fig:violin:PC}
		\includegraphics[width=0.23\textwidth]{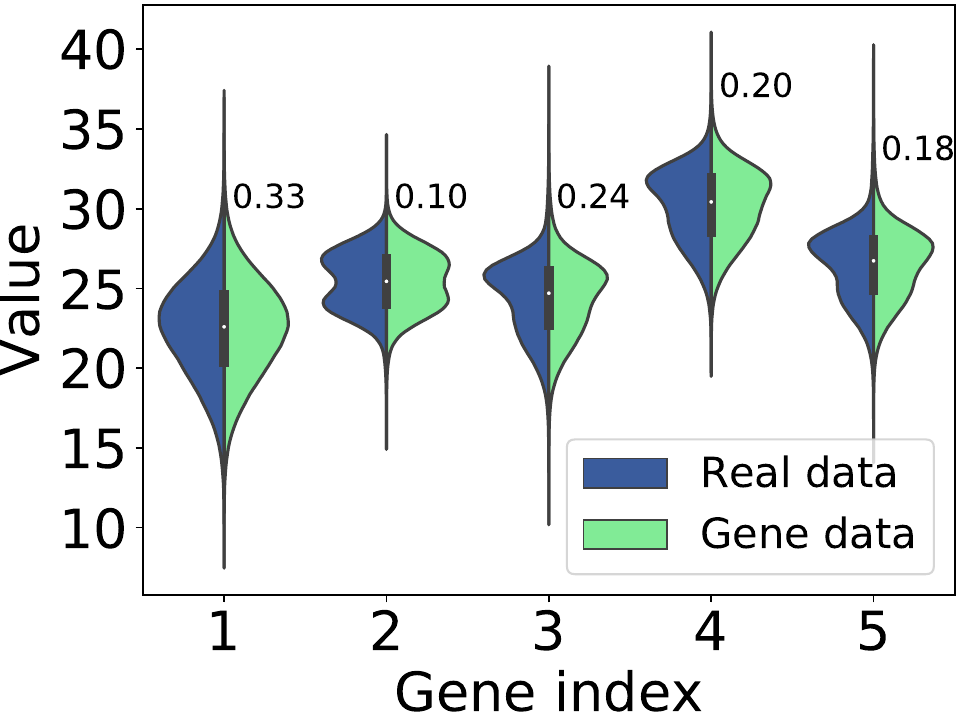}
	}
	\subfigure[\methodshort]{
		\label{fig:violin:PD}
		\includegraphics[width=0.23\textwidth]{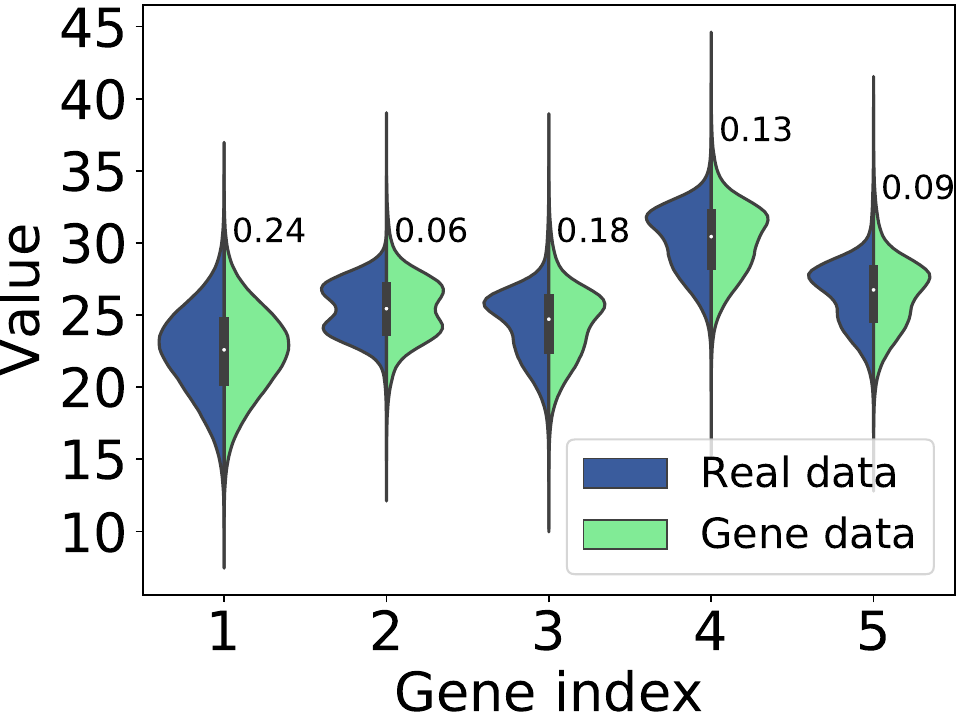}
	}
	
	\caption{The generative performance of our method and baselines on synthetic data by violin plot.  A violin plot is a box plot with a rotated kernel density on each side at different values, making it suitable for visualizing sample distribution. The Y-axis is the sample's value scope and the curve presents the frequency. The X-axis is the gene ID. The floating number represents the KL divergence value between the real and generated data. (a) is the origin distribution of synthetic data. (b), (c) and (d) show the result of CVAE\citep{sohn2015learning}, CGAN\citep{odena2016conditional} and our \methodshort~ method. }
	\label{fig:process}
\end{figure*}
\begin{table*}[thb]
	\centering
	\renewcommand\arraystretch{1.}
	\addtolength{\tabcolsep}{-1pt}
	\begin{tabular}{l|l|cccccccc|cc}
		\toprule
		\multicolumn{2}{c|}{\textbf{\diagbox{Dataset}{Method}}} & \textbf{NN-ED} & \textbf{NN-DTW} & \textbf{NN-CID} & \textbf{FS} & \textbf{TSF} & \textbf{SAX-VSM} & \textbf{MC-DCNN} & \textbf{LSTM} & \textbf{CVAE} & \textbf{\methodshort} \\
		\midrule
		
		\multirow{3}{*}{\textbf{\kaggletrafficshort }} & {Precision} &&&&&&&&&&\\
		~ & {Recall} &&&&&&&&&&\\
		~ & {$F_1$} &&&&&&&&&& \\
		\midrule
		\multirow{3}{*}{\textbf{\timeshort }} & {Precision} & 59.90 & 60.17 & 57.12 & 54.34 & 76.80 & 65.12 & 78.94 & 79.69 & 77.92 & \textbf{80.33} \\
		~ & {Recall} & 34.82 & 41.41 & 40.86 & 43.54 & 52.61 & \textbf{59.96} & 49.27 & 53.56 & 54.12 & 58.17 \\
		~ & {$F_1$} & 44.01 & 49.04 & 47.55 & 48.34 & 62.50 & 62.44 & 60.70 & 64.10 & 64.32 & \textbf{67.45} \\
		\midrule
		\multirow{3}{*}{\textbf{\flowshort }} & {Precision} & 28.51 & 27.14 & 52.65 & 31.66 & 48.11 & 62.71 & 53.77 & 60.25 & 63.27 & \textbf{71.50} \\
		~ & {Recall} & 19.33 & 21.73 & 10.25 &16.73 & 21.04 & 28.41 & 5.79 & 28.01 & 26.78 & \textbf{33.15} \\
		~ & {$F_1$} & 23.01 & 24.13 & 17.05 & 21.84 & 29.13 & 40.11 & 10.38 & 38.23 & 37.57 & \textbf{45.34} \\
		\midrule
		\multirow{3}{*}{\textbf{\trafficshort }} & {Precision} & 54.43 & 51.95 & 56.12 & 65.17 & 54.20 & 72.22 & 76.79 & 56.21 & 74.86 & \textbf{80.23} \\
		~ & {Recall} & 47.88 & 52.43 & 49.26 & 58.82 & 60.94 & 59.05 & \textbf{66.13} & 53.15 & 59.22 & 64.57 \\
		~ & {$F_1$} & 50.95 & 52.14 & 52.44 & 61.85 & 57.42 & 64.94 & 71.06 & 54.63 & 66.14 & \textbf{71.55} \\
		\bottomrule 
	\end{tabular}
	\caption{Classification performance on five real datasets with different methods (\%).The \textbf{bold} indicates the best performance of all the methods and parentheses indicate the number of gene $K$ with the best performance in \secref{sec:exp:pa}.}
	\label{tb:exp:classresult}
\end{table*}
\subsection{Datasets}
We employ five datasets to construct our experiments, including a synthetic dataset and four real-world datasets. 
A synthetic dataset is used to validate the recognition and generation of genes, and real-world datasets are used to validate the effectiveness of \methodshort~by application.
One real-world dataset comes from \textit{Kaggle}\footnote{https://www.kaggle.com}.
The State Grid of China, the largest utility company in the world, and China Telecom, the major mobile service provider in China, provide the other three datasets. 

\vpara{Synthetic.}
We generate five clusters of synthetic samples in $ \mathbb{R}^{N \times T \times S}$. Each sample is a multivariate series with 10 sequential windows; each segment has 20 time points, and each point contains 3 variables. Each cluster has 10K samples.
In particular, for the $k$-th cluster, each dimension of a sample is generated using a mixed Gaussian distribution with mean $\mu$ and standard deviation $\sigma$:  $X_k\sim N(\mu_{k1}, \sigma_{k1}^2)+N(\mu_{k2}, \sigma_{k2}^2)$. The mean $\mu$ and standard deviation $\sigma$ are acquired randomly, $\mu \in [20, 30], \sigma \in [0, 5]$

\vpara{\kaggletraffic (\kaggletrafficshort).}
This dataset comes from Kaggle, which is taken from Jul 1st 2015 up until Dec 31st 2016 and each data point is the number of daily views of the Wikipedia article.
We set a classification task of predicting whether there will be rapid growth (the curve slope greater than 1) in next months (30 days) based on the most recent readings in the past year (12 months). In total, we extract 105k negative cases and 38k positive cases from 145k daily readings. 
 
\vpara{\flowdataset(\flowshort).} 
This dataset is provided by China Telecom. It consists of around 242K network flow series, each of which describes hourly in- and out-flow of different servers, spanning from Apr 1st 2017 to May 10th 2017.  
When an abnormal flow goes through server ports, the alarm states will be recorded.
Our goal is to use the daily network flow data within 15 days to predict if there will be an abnormal flow in the next day. 
In total, we identify 2K abnormal flow series and 240K normal ones.
 
\vpara{\trafficdataset (\trafficshort).} 
This dataset is also provided by China Telecom. 
It includes daily mobile traffic usage for 120K users from Aug. 1st 2017 to Nov. 30th 2017.  
For a user in each day, we obtain 12 kinds of traffic usage records (e.g., total usage, local usage, etc.).  
In this case, we predict whether a user will switch to a new monthly plan, which is associated with high limitation of mobile traffic, according to her recent three-month traffic usage.
Considering only 0.05\% of all users adopt the new plan, 
we use an under-sampling method and obtain a balanced data subset with 16K instances for cross-validation. 

\vpara{\timedataset(\timeshort).} 
This dataset is provided by the State Grid of China. It consists of around 4 million clock error series, each of which describes the deviation time, compared with the standard time, and the communication delay of different watt-hour meters per week, The duration is from Feb. 2016 to Feb. 2018. When the deviation time exceeds 120s, the meter will be marked as abnormal. Our goal is to predict the potential abnormal watt-hour meters in the next month by utilizing clock data from the past 12 months.
In total, we identify 0.5 million abnormal clock error series and 3.5 million normal ones. We will give a more concrete description of the background of this dataset in Section~\ref{sec:interpreter}.

\subsection{Validation on Synthetic Data}
\begin{table}[h]
	\centering
	\renewcommand\arraystretch{1.1}
	\addtolength{\tabcolsep}{-2pt}
	\caption{Recognition performance on Synthetic data}
	\begin{tabular}{l|cccccc}
		\hline
		Metric  & K-means & Agglo & Birch & HMM & GMM & \methodshort \\ 
		\hline
		Homogeneity & 0.546 & 0.533 & 0.537 & 0.612 & 0.637 & \textbf{0.674} \\
		Silhouette score & 0.091 & 0.089 & 0.092 & 0.101 & 0.112 & \textbf{0.158} \\
		\hline 
	\end{tabular}
	\normalsize
	\label{tb:exp:assign}
\end{table}
\vpara{Performance on gene recognition.}
In the synthetic data, we set supervised (homogeneity) and unsupervised (silhouette coefficient) evaluation metrics. The homogeneity score indicates whether all of its subsets contain only data points which are members of a single gene, and the silhouette score indicates how well each object lies within its gene.
We compare \methodshort's result with those obtained by several different clustering algorithms, including K-means clustering, Agglomerative, Birch clustering, Hidden Markov Model (HMM)~\citep{Yang2014HMM} and Gaussian Mixture Model (GMM)~\citep{bouttefroy2010on} . 
As \tableref{tb:exp:assign} shows, K-means performs relatively better than Agglomerative, Birch clustering, which illustrates the distance is a significant indicator for the high-dimensional time series. The performance of HMM and GMM presents that distribution is critical for modeling time series. \methodshort achieves the highest score in both homogeneity and silhouette score, which suggests that classification network $\mathbf{C}$ captures the sequential patterns in segments and outperforms in distinguishing genes.

\vpara{Performance on gene generation.}
\figref{fig:process} presents the generative distribution of each gene learned by different methods on synthetic data.
According to the result of CVAE (\figref{fig:process}(b)), each generated sample shows a similar mean but different variance. 
CGAN's generated samples are similar to real ones (\figref{fig:process}(c)), and can even fit bimodal distribution as the second gene.
We can see that \methodshort~obtains better results than CGAN and CVAE, as is more similar to the distributions of original samples.
This proves that \methodshort~performs better at capturing the distribution patterns of segments.

\subsection{Predicting Future Event}
We then evaluate our proposed model in terms of its accuracy in predicting future events, which then turns into a classification problem of $\mathcal{Y} = \pi$ given $\mathcal{X}$. 
We compare our proposed model against the following night baseline models, which have proven to be competitive across a wide variety of prediction tasks:
\begin{itemize} [leftmargin=*]
	\item \textit{NN-ED, NN-DTW and NN-CID:}  
	Given a sample, these methods calculate their nearest neighbor in the training data and use the nearest neighbor's label to classify the given sample. 
	To quantify the distance between samples, they consider different metrics, which are, respectively, Euclidean Distance, Dynamic Time Warping~\citep{berndt1994using} and Complexity Invariant Distance~\citep{batista2011complexity}. 
	\item \textit{Fast Shapelets (FS):} This is a fast shapelets algorithm that uses shapelets as features for classification~\citep{rakthanmanon2013fast}. 
	\item \textit{Time Series Forest (TSF):} This is a tree-ensemble method that derives features from the intervals of each series~\citep{deng2013time}. 
	\item \textit{SAX-VSM:} This is a dictionary method that derives features from the intervals of each series~\citep{senin2013sax-vsm}. 
	\item \textit{MC-DCNN and LSTM:}  These are two deep neural network-based methods proposed in~\citep{zheng2014time} and~\citep{hochreiter1997long} respectively. 
\end{itemize}
Besides the above methods, we further consider the following generative models as baselines:
\begin{itemize} [leftmargin=*]
	\item \textit{CVAE:} This method uses CVAE~\citep{sohn2015learning} as gene $\mathbf{G}$ without discriminator  and uses the same feature fusion method for prediction.
	\item \textit{\methodshort:} This is the proposed method. We use $\mathcal{L}_{event}$ as $\mathcal{L}_{app}$ to train \methodshort networks.
\end{itemize}
\vpara{Comparison results.} \tableref{tb:exp:classresult} compares the results of event prediction. 
We use precision, recall and F-measures ($F_{0.5}$) as metrics.
Here, we prefer to use $F_{0.5}$ as metric 
because precision is  more important than recall in this scene.
We observe that all quantifying-distance methods based on nearest neighbors perform similarly but are unstable,  which may be attributed to peculiarities in the data, since the NN-DTW method does not outperform on the INS and TMP datasets. 
Moreover, feature-extracted methods have relatively better recall on MCE and TMP datasets, such as the dictionary-method SAX-VSM, but precisions do not outperform simultaneously,
which may not adapt to the unbalanced sample. 
The neural network approaches (MC-DCNN, LSTM) perform poorly on small-scale data (Earthquakes), for they might be more suitable for processing large-scale data due to their model complexity.
The generative method utilizes the genes' distribution patterns and models the behavior evolution, which leads to a better performance on the five real-world datasets. CVAE outperforms near-neighbor methods on all datasets, which attributes to modeling behavior evolution behind the time series. 
As we expected, due to the ability to fitting distribution better, \methodshort performs better than CVAE and outperforms these baselines.

\subsection{Parameter Analysis}
\label{sec:exp:pa}
\begin{figure}[t]
	\centering
	\begin{minipage}{0.5\textwidth}
		\includegraphics[width=1.\textwidth]{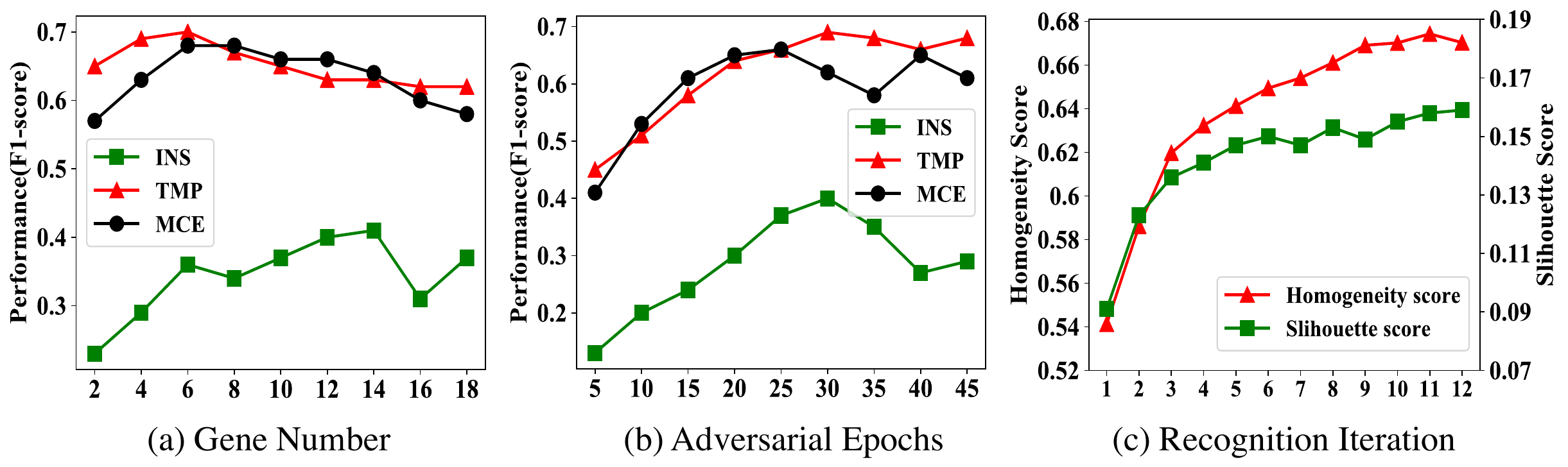}
		\caption{Model parameter analysis.  (a), (b) testing on real datasets and (c) testing on a synthetic one. (a) presents the sensitivity of the genes' number. (b) shows the sensitivity of the adversarial training epochs. (c) shows the sensitivity of the iterations of gene recognition. 
		}
		\label{fig:para}
	\end{minipage}
\end{figure}
Finally, we study the sensitivity of the model parameters: iterations of recognition ($\mathbf{C}$), adversarial epochs ($\mathbf{E}, \mathbf{G}, \mathbf{D}$) and the number of genes ($K$). 
We present the results on synthetic dataset and three real-world datasets. we use the performance of F1 score, which is based on the future event prediction, as metric, and compare the different hyper-parameters, 
\figref{fig:para}(a) shows that the gene number $K$ influences the model performance differently on the three real-world datasets. 
The F1 score is not bound to improve as the gene number increases and the peaks of gene number in TMP and MCE datasets are around 6 and 8 but the peak in INS dataset is around 14. 
We conclude that this is an empirically determined parameter that may vary across different datasets.
\figref{fig:para}(b) presents that the performance of \methodshort on future event prediction is positively related to the training epoch at first, after which there are fluctuations that may be caused by the instability of adversarial training. As shown in \figref{fig:para}(b), the best parameter of adversarial training epochs in the three real-world datasets are around 25 to 30. Finally, \figref{fig:para}(c) shows how $\mathbf{C}$ influences the performance of gene recognition. We compare the homogeneity score and silhouette score in different iterations.  We can see the fully trained classifier is the prerequisite for learning patterns of the gene. The growth curve approximates the log function, which grows fast in the early stage and tends to stabilize in the later stage.

%% file: interpreter.tex

\section{Application}
\label{sec:interpreter}
\begin{figure*}[h]
	\centering
	\includegraphics[width=1.\textwidth]{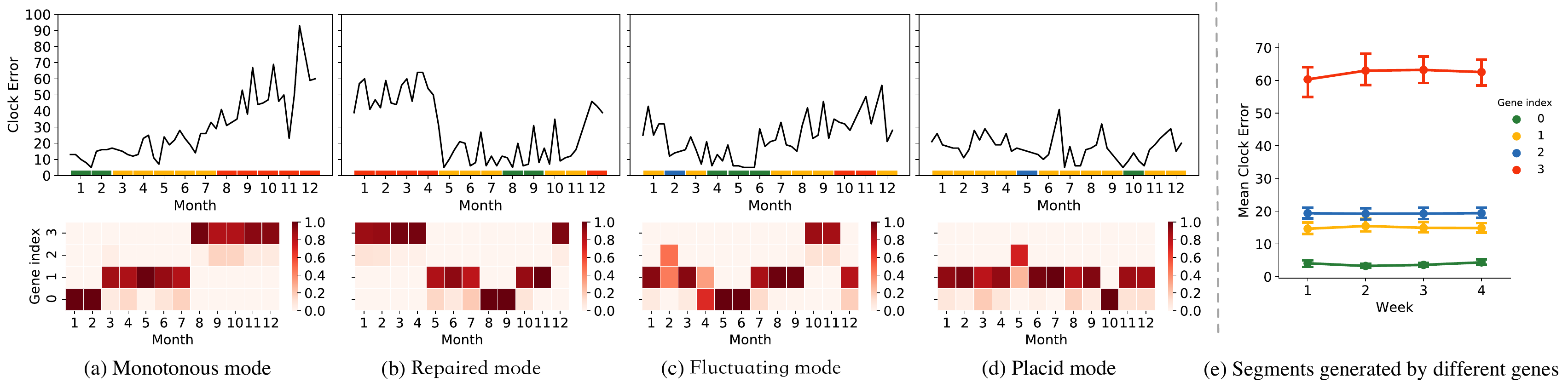}
	
	\caption{A real-world application of \methodshort based on the dataset provided by the State Grid of China. In the figure, (a), (b), (c) and (d) present four different types of time series about watt-hour meters' clock error. The upper figure shows raw curve of clock error and the marks of the genes, while the lower figure visualizes the behavior sequence $\mathcal{A}$ of genes computed by \methodshort.
	These figures illustrate four different evolution modes (i.e., monotonous, repaired, fluctuating, placid) of watt-hour meters. (e) shows the average segments 
	generated from four different genes. 
 }
	\label{fig:interpretation}
\end{figure*}

We have deployed \methodshort ~to State Grid Wenzhou Power Supply Co. Ltd. to detect abnormal status of watt-hour meters.
More specifically, \methodshort ~will detect high-risk meters at the beginning of every month, identify the factor that causes the abnormality by analyzing the behavior evolution of meters (Here, the behaviors of watt-meters are the different levels of indications), and suggest engineers to adopt corresponding strategies in advance.  
It turns out that \methodshort ~is able to reduce the maintenance workloads of watt-hour meters by 50\%, which costs around \$300 million per year previously.
In this section, we will introduce the background of this application and present a case study to demonstrate that \methodshort ~not only achieves around 80\% precision of anomaly prediction, but precisely captures the different evolution modes of watt-hour meters. 
For simplicity, we use four genes to present this application.

\vpara{Background.}
In a watt-hour meters, the clock is one of the basic and the most important components, whose accuracy is directly related to whether the meter can accurately measure the data in different time periods. 
However, due to several factors, such as inaccurate clock synchronization signals, the crystal oscillator of device, communication delay, and device response delay, the time recorded by the watt-hour meter may \textit{deviate} from the standard time inevitably. 
Furthermore, different factors on the watt-hour meter will lead the clock error to evolve by following different \textit{modes}. 
For example, the crystal oscillator will cause the clock error to fluctuate in one direction, while unstable communication environment will lead to the swinging clock error. 
Therefore, discovering these different evolution modes of clock errors has great significance for diagnosing and maintaining watt-hour meters.
Our method is expected to not only predict the error state of the given watt-hour meter, but also reveals different evolution modes of clock errors. 
In particular, we manually find four most representative evolution modes as follows: 
\begin{itemize} [leftmargin=*]
	\item Monotonous mode: The clock error fluctuates in one direction over time (12 months), which may be caused by the crystal oscillator of device.
	\item Repaired mode: The clock error will recover at a certain time, which may caused by receiving the clock synchronization signals from the superior terminal.
	\item Fluctuating mode: The clock error fluctuates violently, which may be caused by the poor communication environment.
	\item Placid mode: The clock error fluctuates gently, which is the ideal status of healthy watt-hour meters.
\end{itemize}
The above four patterns have covered over 93\% samples. 
Therefore we mainly study these representative patterns and ignore others (e.g., sudden drop or rise of clock error) in this section. 

\vpara{Recognizing evolution modes.} 
Is the proposed model able to disclose and model these four evolution modes? 
Before we answer this question, we present the different watt-meters' behaviors by the average value of clock error that generated by different genes in \figref{fig:interpretation}(e). 
We see that average clock error of gene \#3 is significant larger than that of other genes, which suggests that gene \#3 denotes an ``abnormal behavior'' corresponding to abnormal watt-hour meters. 

\figref{fig:interpretation}(a)-(d) visualizes four watt-hour meters with observed clock errors that follow different evolution modes (in plots) and how \methodshort recognizes genes to each segment (in heat map, where the y-axis indicates the probability of each gene being recognized to the segments at different time). 
For example, the clock error that evolves by following the monotonous mode keeps small value at first, and will keep growing over time (\figref{fig:interpretation}(a)). 
Correspondingly, we see that our model captures this process and tends to recognize ``normal behavior'' to the sample first, while eventually determines it has the ``abnormal behavior'' (i.e., gene \#3). 
Therefore, we see that the way our model learn genes is identical to the monotonous mode. 
Similar results can be observed in other three modes. 
In particular, our model recognizes ``normal behaviors'' and ``abnormal behaviors'' alternately to the watt-hour meter with repaired mode and fluctuating mode (\figref{fig:interpretation}(b)-(c)), while tends to keep recognizing ``normal behaviors'' to the samples with placid mode (\figref{fig:interpretation}(d)).  

%% file: related.tex

\section{Related Work}
\vpara{Time series modeling.} Time series modeling have been used in many domains,
such as 
anomaly detection (e.g., abnormal mutation \citep{chapfuwa2018adversarial} and gradual decline \citep{janakiraman2017finding, Du2016Recurrent}); 
human behavior recognition (e.g., circadian rhythms and cyclic variation \citep{althoff2017harnessing, pierson2018modeling}); 
and biology applications (e.g., the hormonal cycles \citep{chiazze1968the}). 
The majority have concentrated on different distance measurements to model evolutionary data, such as dynamic time warping \citep{Lines2015Time,chiazze1968the}, move–split–merge~\citep{Stefan2013The},
complexity-invariant distance~\citep{Batista2014CID} and elastic ensemble~\citep{Lines2015Time,chapfuwa2018adversarial}. 
Some methods focus on sequence-clustering by distance~\citep{Zhou2017Patient, althoff2017harnessing},
which aims to find a better distance to model series and enhance the clustering performance. However, this is different from our task.
Some feature-based classifiers have also been explored~\citep{baydogan2016time, Kurashima2018Modeling}, which are distinguished by the frequency of segment repetition rather than by its distribution. They form frequency counts of the recurring patterns, then build classifiers based on the resulting histograms~\citep{Lin2012Rotation, Xu2018Unsupervised}. 

Model-based algorithms fit a generative model to each series, then measure the similarity between the series using the similarity of the model's parameters. The parametric approaches used include fitting auto-regressive models\citep{
Shokoohi2015Discovery}, hidden Markov models\citep{Yang2014HMM, Wu2017Retrospective} and kernel models\citep{
Kurashima2018Modeling}, which rely on the artificial knowledges. Recently, many models using neural networks have been proposed \citep{Wang2018Multilevel, Wang2018PredRNN, Bi2018Autoregressive}.
Deep learning methods for series data have mostly been studied in high-level patterns representation. The main idea behind these approaches is that of modeling the fusion of multiple factors like time or space, etc. .

\hide{
\vpara{Event prediction.} Most studies in this area focus on predictions of next events, including unix commands \citep{Davison1998Predicting}, user interface actions to enable interface adaption \citep{Gorniak2000Predicting}, web page requests allowing for pref-etching and latency reduction \citep{Zukerman1999Predicting}.
Many of these works (e.g., \citep{Liu2016Predicting,Rendle2010Factorizing,Kooti2016Portrait,Baeza2015Predicting}) have formulated the problem as a discrete-time sequence prediction task and used Markov models. However, Markov models assume unit time steps and are further unable to capture long-range dependencies since the overall state-space will grow exponentially in the number of time steps considered \citep{Du2016Recurrent}. Other works have used LSTM models \citep{hochreiter1997long}, which also assume discrete time steps and are limited in their interpretability.

In contrast, we also model and predict when the next event will occur, which is critical to recommendations and reminders at the right time. In addition, instead of specific time series ordering, we represent different patterns by their distributions' estimation of each segment in the time window and model the overall evolution in the past.
}

\vpara{Deep generative models.} Generative models have recently attracted significant attention, and the nonparametric learning ability over large (unlabeled) data endows them with more potential and vitality. 
There have been many recent developments of deep generative models \citep{karras2018progressive, chapfuwa2018adversarial, Xu2018Unsupervised, arjovsky2017towards}. Since deep hierarchical architectures allow them to capture complex structures in the data, all these methods show promising results in generating natural sample that are far more realistic than conventional generative models. Among them are two main themes: Variational Auto-encoder (VAE) \citep{Kingma2013Auto} and Generative Adversarial Network (GAN)  \citep{Goodfellow2014Generative}.
Variational Auto-encoder (VAE) pairs a differentiable encoder network with a decoder/generative network. The encoder network intended to represent a data instance in a latent hidden space, which the inference is done via variational methods. A disadvantage of VAE is that, because of the injected noise and imperfect element-wise measures such as the squared error, the generated samples are often blurry \citep{2017arXiv170310155B}.
Generative Adversarial Network (GAN) is another popular generative model. It simultaneously trains two models: a generative model to synthesize samples, and a discriminative model to differentiate between natural and synthesized samples. However, the GAN model is hard to converge in the training stage and the samples generated from GAN are often far from natural. Class conditional synthesis can significantly improve the quality of the generated samples\citep{sohn2015learning, odena2016conditional}. As a result, a lot of recent research has focused on finding better training algorithms \citep{
karras2018progressive} for GANs as well as gaining better theoretically understanding of their training dynamics \citep{arjovsky2017towards, Mescheder2018Which}

Our model differs from all these models. We use a classifier to learn the genes corresponding to segments, then use a CVAE-GAN structure~\citep{2017arXiv170310155B}  to estimate the distribution patterns. We predict the future events and values based on the distribution evolution.

\label{sec:related}

%% file: conclusion.tex

\section{Conclusions}
\label{sec:conclusion}

In this paper, we study the problem of capturing 
the behavior evolution behind the time series and predicting future events. 
Based on that, we define the ``gene'', 
to model the generation of time series from different behaviors.
We take advantage of CVAE-GAN structure to learn the genes and estimate segments' distribution patterns. Additionally, a classifier is learned to select gene for each segment.
We propose \methodname~(\methodshort) that places these two tasks into a uniform framework, which consists of a classifier to learn ``gene'' to different segments, and learning distribution patterns by the adversarial generator. 
We apply these patterns into modeling behavior evolution by a recursive structure.
To validate the effectiveness of the proposed model, we conduct sufficient experiments based on both synthetic and real-world datasets.
Experimental results show that our model outperforms several state-of-the-art baseline methods.
Meanwhile, we demonstrate the interpretability of our model by applying it to the real maintenance of watt-hour meters in the State Grid Corporation of China.